\documentclass[10pt,twocolumn,letterpaper]{article}

\usepackage[pagenumbers]{wacv}

\newcommand{\ourparagraph}[1]{\vspace{3px}\noindent\textbf{#1}}

\usepackage[dvipsnames]{xcolor}

\definecolor{cvprblue}{rgb}{0.21,0.39,0.84}

\usepackage[pagebackref,breaklinks,colorlinks,citecolor=cvprblue]{hyperref}
\usepackage{graphicx}
\usepackage{amsmath}
\usepackage{amssymb}
\usepackage{booktabs}
\usepackage{hyperref}      
\usepackage{url}           
\usepackage{amsfonts}     
\usepackage{nicefrac}    
\usepackage{microtype}   
\usepackage{xcolor}    
\usepackage{amsfonts}
\usepackage{multirow}
\usepackage{enumitem}
\usepackage{array}
\usepackage{algorithm}
\usepackage{algorithmicx}
\usepackage{algpseudocode}
\usepackage{upgreek} 
\usepackage{bm} 
\usepackage{multicol}
\usepackage{colortbl}
\usepackage{booktabs}

\newcommand{\bx}{\mathbf{x}}

\newcommand{\bz}{\mathbf{z}}

\newcommand*{\tran}{^{\mkern-1.5mu\mathsf{T}}}

\usepackage[pagebackref,breaklinks,colorlinks]{hyperref}

\usepackage[capitalize]{cleveref}
\crefname{section}{Sec.}{Secs.}
\Crefname{section}{Section}{Sections}
\Crefname{table}{Table}{Tables}
\crefname{table}{Tab.}{Tabs.}

\def\confName{WACV}

\begin{document}

\title{Disentangling Disentangled Representations:\\Towards Improved Latent Units via Diffusion Models}

\author{
Youngjun Jun \quad 
Jiwoo Park\quad 
Kyobin Choo\quad 
Tae Eun Choi\quad 
Seong Jae Hwang\thanks{Corresponding author}\\ 
Yonsei University
\\{\tt\small \{youngjun, wldn1677, chu, teunchoi, seongjae\}@yonsei.ac.kr}
}

\maketitle

\begin{abstract}
Disentangled representation learning (DRL) aims to break down observed data into core intrinsic factors for a profound understanding of the data. In real-world scenarios, manually defining and labeling these factors are non-trivial, making unsupervised methods attractive. Recently, there have been limited explorations of utilizing diffusion models (DMs), which are already mainstream in generative modeling, for unsupervised DRL. They implement their own inductive bias to ensure that each latent unit input to the DM expresses only one distinct factor. In this context, we design \textbf{Dynamic Gaussian Anchoring} to enforce attribute-separated latent units for more interpretable DRL. This unconventional inductive bias explicitly delineates the decision boundaries between attributes while also promoting the independence among latent units. Additionally, we also propose \textbf{Skip Dropout} technique, which easily modifies the denoising U-Net to be more DRL-friendly, addressing its uncooperative nature with the disentangling feature extractor. Our methods, which carefully consider the latent unit semantics and the distinct DM structure, enhance the practicality of DM-based disentangled representations, demonstrating state-of-the-art disentanglement performance on both synthetic and real data, as well as advantages in downstream tasks.
\end{abstract}

\begin{figure}[t!]
    \centering
    \includegraphics[width=1.0\linewidth]{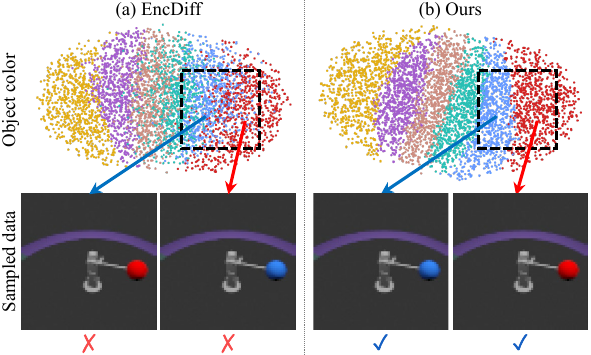}
    \caption{
     Visualization of latent units in EncDiff and Ours. 
     The top figures visualize the latent unit representing \texttt{object color} using the dimensionality reduction method PaCMAP\cite{pacmap} for multiple data points. The bottom figures show the results of conditional generation of data by sampling the latent unit from the \texttt{blue} and \texttt{red} regions, respectively. (a) In EncDiff\cite{encdiff}, the boundary between color regions is ambiguous, so a latent unit representing \texttt{red} can be sampled even in the \texttt{blue} region, and vice versa. (b) Using our proposed method, we achieve an interpretable latent unit by clearly defining the boundaries between attributes, ensuring that only one color appears in each color region.
    }
    \label{fig:motivation}
\end{figure}

\section{Introduction}
\label{sec:itro}
Disentangled representation learning (DRL) aims to uncover the fundamental factors within the observed data \cite{bengio,torward-def,beta-vae,disco}.\let\thefootnote\relax\footnote{\scriptsize{Project Page:~\url{https://youngjun-jun.github.io/dis-dis-rep}}} 
A disentangled representation is a set of latent units where each unit is dependent on only one factor and invariant to the remaining factors\cite{bengio,drl,peters,building,deep-learning}. 
Attaining such representation is essential for understanding data and generalizing models, making it a fundamental goal in the machine learning field \cite{bengio,overlooked}.

To achieve DRL, supervised methods that require annotations of manually-defined factors have been attempted \cite{multi-vae,finegan}. However, in the real world, factors are complex and exist on a non-discrete spectrum (\eg, image brightness), making annotation challenging. Thus, unsupervised learning methods are extensively being explored \cite{disco,dava,fdae}. They commonly utilize the latent space of generative models, which capture the semantic information of images. Meanwhile, it is known that achieving disentanglement in an unsupervised manner is impossible without explicit inductive biases \cite{challenging}, such as regularization for statistical independence and architectural modeling \cite{tripod,disco,disdiff}. Thus, the essence of unsupervised DRL lies in how effectively a suitable inductive bias is implemented into the target generative model.

Recently, diffusion models (DMs) \cite{ddpm,ldm,edm2} have emerged as a cornerstone of generative models, known for their superior generation quality and stability. In particular, DMs not only effectively encode rich information about the input image, but also ensure that the manipulated features in the latent space are directly reflected in the generated image \cite{asyrp}. This suggests that DMs are attractive target generative models for unsupervised DRL \cite{disdiff,fdae,encdiff}. Naturally, existing DM-based DRL methods focus on providing appropriate inductive biases to DMs to achieve independence among latent units \cite{disdiff,fdae}. Meanwhile, the significance of each latent unit accurately reflecting its corresponding factor has not yet been considered. Therefore, to enhance the practical usefulness and interpretability of representations, we not only focus on the independence of the factors (\eg, \texttt{object color}) represented by the latent units, but also on how faithfully each latent unit reflects the attributes (\eg, \texttt{red}, \texttt{blue}) of those factors.

To understand this, let us consider a scenario highlighting the necessity of attribute-separated latent units. \cref{fig:motivation} visualizes a single latent unit (responsible for \texttt{object color}) across multiple data points using both the DM-based state-of-the-art (SOTA) DRL method EncDiff \cite{encdiff} and our proposed method. In \cref{fig:motivation}a, the data points for EncDiff are not clearly separated into distinct attributes (\eg, \texttt{red}, \texttt{blue}). This entanglement might actually result in selecting a \texttt{blue} image when sampling from the \texttt{red} region. Such representation with semantically ambiguous latent units cannot guarantee the intended downstream uses (\eg, manipulating \texttt{object color} factor of intended color). In contrast, as shown in \cref{fig:motivation}b, ours exhibits clear decision boundaries between attributes, ensuring that images sampled from each color region faithfully reflect the intended color. This enhances interpretability by indicating which intrinsic factor a latent unit represents in real-world data where labels are absent, suggesting a more practical representation. Moreover, latent units with a high association (\ie, mutual information) with a specific factor naturally become less dependent on other factors, thereby promoting disentanglement.

Thus, we first propose \textbf{Dynamic Gaussian Anchoring (DyGA)}, an inductive bias that clarifies the decision boundaries between attributes of a latent unit in DM-based DRL. DyGA dynamically selects anchors for attribute clusters in the latent space and shifts ambiguous points at the cluster boundaries toward these anchors. This well-organized attribute latent unit is used as a condition for the DM, learning to simultaneously perform disentanglement of the feature extractor and image generation of the DM.

However, since DM can also be trained unconditionally, it may ignore unstable latent units during early training and rely less on the feature extractor. This is due to the peculiarity of DM structures which receive the latent unit only as an auxiliary condition via U-Net's cross-attention; a new aspect that does not need to be considered in variational autoencoder-based DRL methods \cite{beta-vae,shapes3d-factorvae} where the latent unit is the sole input condition. Ideally, for the diffusion denoising U-Net and feature extractor to learn complementarily, the training should be guided so that, rather than the noisy image input, the latent unit input determines the core image elements.

Therefore, we also propose \textbf{Skip Dropout (SD)}, an effective technique that make DM networks more DRL-friendly with a simple adjustment. SD drops U-Net's skip connection features from the noisy image input, ensuring that the DM training focuses on the latent unit features and the feature extractor, which are the key to disentanglement. In conclusion, through comprehensive modeling for both the latent unit semantics and DM structure, our proposed methods present the potential of DM for more interpretable disentanglement.

\vspace{3px}
\noindent\textbf{Contributions.} Our main contributions are as follows:
\begin{itemize}[noitemsep, nolistsep, leftmargin=*]
    \item We design \textit{Dynamic Gaussian Anchoring}, a novel inductive bias for interpretable disentanglement that guides attribute-separated latent units through anchoring-based manipulation in the latent space.
    \item We propose \textit{Skip Dropout}, a modification of the U-Net architecture to enhance the disentangling functionality of the feature extractor within the DM-based DRL framework.
    \item Our techniques have been applied to existing DM-based methods, achieving SOTA unsupervised DRL performance, and the obtained representations demonstrate strengths in downstream tasks as well.
\end{itemize}

\begin{figure*}[t!]
    \centering
    \includegraphics[width=1.0\textwidth]{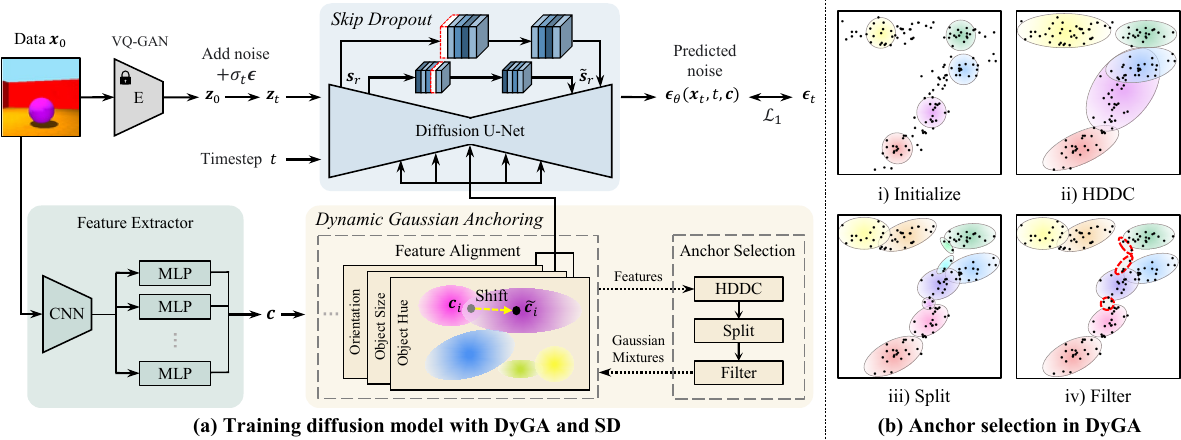}
    \caption{
    Training framework with proposed methods. (a) During the diffusion model training, the features generated by the feature extractor are shifted towards the mean direction of the Gaussian for each feature unit based on the selected anchor, becoming the condition for the diffusion model. To ensure the diffusion U-Net effectively utilizes the conditions created by the feature extractor, a skip dropout strategy is employed. (b) The process of anchoring Gaussian distributions involves: i) initializing the Gaussian mixture, ii) performing HDDC using the EM algorithm, iii) adjusting the number of Gaussians by splitting them according to criteria, and iv) filtering out unnecessary Gaussians.
    }
    \label{fig:model}
\end{figure*}

\section{Related Work}
\vspace{-5pt}
\label{sec:rela}

\noindent\textbf{Disentangled Representation Learning.} 
Disentangled representation learning (DRL) aims to identify underlying factors from observable data, with variational autoencoders (VAEs) initially favored as generative models because their decoders generate exclusively using the information-rich latent space.
$\beta$-VAE \cite{beta-vae} and AnnealVAE \cite{anneal-vae}, for instance, use regularization and information bottleneck principles for this purpose. InfoVAE \cite{info-vae} and FactorVAE \cite{shapes3d-factorvae} focus on mutual information and total correlation, respectively. However, such regularizations alone are insufficient for unsupervised DRL, as explicit inductive bias has been proven to be mandatory for both models and data sets \cite{challenging}.
In response, QLAE \cite{infomec-lq} introduces learnable latent quantization to promote an organized latent space, providing an inductive bias for meaningful and consistent representations. Tripod \cite{tripod} suggests a quantization method with finite scalar codebooks after identifying the limitations of latent quantization learning \cite{fsq}. Despite these efforts, VAE-based models face a trade-off between image quality and disentanglement \cite{beta-vae, mig-betaTCVAE, shapes3d-factorvae}. Consequently, InfoGAN \cite{info-gan} demonstrated that generative adversarial networks (GANs) \cite{gan} have informative latent spaces and can achieve disentanglement via the lower bounds of mutual information. Pre-trained GAN-based methods such as LD \cite{LD}, CF \cite{CF}, GS \cite{GS}, and DS \cite{DS} have also been explored. However, GANs' instability and mode collapse issues limit their use, leading to recent studies on unsupervised disentanglement with diffusion models \cite{disdiff, fdae, encdiff}.

\ourparagraph{Diffusion Model-based DRL.}
Due to their generative performance and rich latent space \cite{asyrp}, diffusion models have recently gained attention for disentangling their latent space.
DisDiff \cite{disdiff} attempted to obtain independent latent vectors using CLUB \cite{club} of mutual information, while FDAE \cite{fdae} aimed to explicitly improve interpretability by forcing the latent space into separate content and mask codes. Subsequently, EncDiff \cite{encdiff} tried to disentangle the latent space by emphasizing that cross-attention can provide inductive bias through an information bottleneck using only diffusion loss.
The general framework, detailed next, establishes a separate feature extractor to derive disentangled representations alongside diffusion model training. However, little has been done to analyze these representations to determine if they truly capture the underlying attributes of each factor. This work delves deeper, introducing a simple yet effective inductive bias for this purpose.

\section{Methods}
\label{sec:method}

In this section, we begin with a brief explanation of diffusion models. Next, we introduce the overall framework (Sec.~\ref{sec:framework}), propose our methodological contributions: (1) Dynamic Gaussian Anchoring as a new inductive bias for diffusion models (Sec.~\ref{sec:DyGA}), and (2) skip dropout to enhance the training of the feature extractor (Sec.~\ref{sec:skip}).

\vspace{3px}
\noindent\textbf{Diffusion Models.}
Diffusion models\cite{ddpm,ldm,edm} are a type of latent variable model that reconstruct $\bx_0 \sim p_{\text{data}}(\bx_0)$ from $\bx_T \sim \mathcal{N}(\mathbf{0}, \mathbf{I})$ with the following formulation $p_\theta(\bx_0) := \int p_\theta(\bx_{0:T}) \, d\bx_{1:T}$.
Diffusion models use a Markov process known as the forward process, which gradually adds noise to the image through a variance schedule $\beta_1, \dots, \beta_T$ until it becomes $\bx_T \sim \mathcal{N}(\mathbf{0}, \mathbf{I})$:

\small
\begin{equation}
    q(\bx_t \mid \bx_{t-1}) = \mathcal{N}(\bx_t; \sqrt{1 - \beta_t}\bx_{t-1}, \beta_t\mathbf{I}).
\end{equation}
\normalsize
Then, with $\Bar{\alpha_t} = \prod_{s=1}^{t} \alpha_s$ and $\alpha_t = 1 - \beta_t$, the latent variable $\bx_t$ at timestep $t$ can be obtained through the following interpolation $\bx_t = \sqrt{\Bar{\alpha_t}}\bx_0 + \sqrt{1 - \Bar{\alpha_t}}\mathbf{\epsilon}$, where $\mathbf{\epsilon} \sim \mathcal{N}(\mathbf{0}, \mathbf{I})$.
This diffusion model optimizes the network according to the following objective:
\small
\begin{equation}
    \mathcal{L}_\eta = \mathbb{E}_{\bx_0,\mathbf{\epsilon},t}[\eta_t||\mathbf{\epsilon}_\theta(\bx_t,t,\mathbf{c})-\mathbf{\epsilon})||],
\end{equation}  
\normalsize
where $\eta_t$ is the coefficient according to the noise schedule, $\epsilon_\theta(\bx_t, t, \mathbf{c})$ is the predicted noise, and $\mathbf{c}$ is some condition.

\subsection{Diffusion Model Framework for DRL}
\label{sec:framework}

We now introduce our full framework as shown in \cref{fig:model}. Let us first describe the DRL framework involving diffusion models (DM) which is commonly used in recent DM-based DRL methods \cite{fdae,encdiff}. Hence, our technical contributions in Sec.~\ref{sec:DyGA} and Sec.~\ref{sec:skip} can be readily applied to methods based on this framework.

\vspace{3px}
\noindent\textbf{Latent Diffusion Model.}
Latent diffusion models (LDM) \cite{ldm} are a type of diffusion model that significantly reduces computational cost by training the diffusion model on images compressed in the latent space rather than the pixel space. LDMs have gained widespread recognition for their utility \cite{sdxl,dreambooth,controlnet}. 
We adopt the widely-used LDM framework with VQ-GAN in order to train conditional generation by conditioning the latent units through cross-attention. 

\vspace{3px}
\noindent\textbf{Feature Extractor.}
The feature extractor extracts a compressed feature from the image, consisting of $N$ latent units. Each latent unit is trained to represent one of the intrinsic factors constituting the image. Similar to previous research \cite{encdiff}, the structure of the feature extractor includes a simple CNN that extracts semantic information from the image and an MLP layer applied to each latent unit. Each latent unit is composed of a $D$-dimensional vector, set to $D=32$ following prior studies \cite{disco, disdiff}. The entire feature $\mathbf{c}=[\mathbf{c}_1,\dots,\mathbf{c}_N]$, composed of latent units, serves as the condition for the LDM. See the supplement for details.

\subsection{Dynamic Gaussian Anchoring}
\label{sec:DyGA}

In this subsection, we introduce Dynamic Gaussian Anchoring (DyGA) to ensure that each latent unit of the feature $\mathbf{c}$ faithfully reflects each factor. DyGA is divided into two processes. First, \textit{anchor selection} involves determining the anchors based on the features, with the number of anchors being decided dynamically. Second, \textit{feature alignment} delineates the boundaries between attributes represented by latent units by adjusting the features towards the direction of the selected anchor.

\subsubsection{Anchor Selection}
As shown in \cref{fig:model}b, anchor selection involves 1) initializing multivariate Gaussian distributions, 2) fitting the Gaussian mixture via high-dimensional data clustering (HDDC) \cite{hddc}, 3) splitting this Gaussians to dynamically increase the number of Gaussians, and 4) filtering out unnecessary Gaussians. At this point, the anchors become the means of the Gaussians, and the spliting and filtering processes serve to dynamically adapt the number of anchors. Meanwhile, HDDC uses the Expectation-Maximization (EM) algorithm to maximize the likelihood function, which is generally non-convex and has many stationary points\cite{likelihood-func}. This means that HDDC can get trapped in a sub-optimal stationary point. Adjusting the number of anchors provides an opportunity to escape from these stationary points
, as it slightly alters the optimization problem
.
Moreover, it is suitable for unsupervised learning where prior knowledge of the data is unavailable, making it appropriate for cases where the number of attributes within the factors is unknown.

\vspace{3px}
\noindent\textbf{High-dimensional Data Clustering.}
Finite mixture models \cite{Finite-mixture-models} predict data distribution as a Gaussian mixture by maximizing the likelihood function through the Expectation-Maximization (EM) algorithm. However, due to the curse of dimensionality \cite{bellman}, this method is difficult to apply to high-dimensional data. To address this, high-dimensional data clustering (HDDC) \cite{hddc} reduces the high-dimensional problem to a lower-dimensional subspace where the EM update is performed. For Gaussian $\mathcal{N}(\bm{\upmu}_i, \boldsymbol{\Sigma}_i)$, for $i=1,\dots,K$, $\bm{\upmu}_i \in \mathbb{R}^{d}$, and $\boldsymbol{\Sigma}_i \in \mathbb{R}^{d\times d}$, the class conditional covariance matrix $\boldsymbol{\Delta}_i$ is defined as follows:
\small
\begin{equation}
    \boldsymbol{\Delta}_i = \boldsymbol{Q}_i\tran\boldsymbol{\Sigma}_i\boldsymbol{Q}_i,
\end{equation}
\normalsize
where $\boldsymbol{Q}_i$ is the orthogonal matrix of eigenvectors of $\boldsymbol{\Sigma}_i$. Consequently, $\boldsymbol{\Delta}_i$ becomes a diagonal matrix with the eigenvalues of $\boldsymbol{\Sigma}_i$ as its diagonal entries. Among these, we let $d_i$ diagonal entries remain unchanged while the remaining $d-d_i$ entries are tied as a single parameter. We can then define the subspace $\mathbb{F}_i$ spanned by $d_i$ eigenvectors and its orthogonal complement $\mathbb{F}_i^\perp \in \mathbb{R}^{d-d_i}$ such that $\mathbb{F}_i \oplus \mathbb{F}_i^\perp = \mathbb{R}^d$.
Now, we are able to handle high-dimensional data through a Gaussian mixture model in the subspace $\mathbb{F}_i$.
However, HDDC requires the number of Gaussians to be specified in advance and this number remains fixed. We propose two methods to adjust this dynamically.

\ourparagraph{Dynamic Adjustment of the Number of Anchors.}
Gaussians fitted to the features via HDDC represent a stationary point of the likelihood function maximized by the EM algorithm. To address the fact that the optimality of this stationary point cannot be guaranteed\cite{likelihood-func}, there have been attempts to complement this with split/merge strategies in Gaussian mixture models \cite{split-merge-0,split-merge-1,split-merge-2}. However, such attempts have not been explored for high-dimensional data. Additionally, a na\"ive merging strategy may not be suitable for feature alignment, especially when dealing with real-world data where the label may be a continuous value. Therefore, a strategy that dynamically adjusts the number of Gaussians to handle continuous variables is needed.

\vspace{3px}
\noindent\textbf{Gaussian Splitting and Filtering.}
First, Gaussian splitting is based on the density of the Gaussians according to the responsibility, and the responsibility $\upgamma_{ij}$ of the $j$-th Gaussian for feature $\bx_i$ is given by the following equation:
\small
\begin{equation}
    \upgamma_{ij} = \frac{\pi_j \mathcal{N}(\bx_i | \bm{\upmu}_j, \mathbf{\Sigma}_j)}{\sum_{k=1}^{K} \pi_k \mathcal{N}(\bx_i | \bm{\upmu}_k, \mathbf{\Sigma}_k)}
\label{eq:responsibility}
\end{equation}
\normalsize
where $\pi_j$ is the weight of component $j$, $\mathcal{N}(\bx_i | \bm{\upmu}_j, \mathbf{\Sigma}_j)$ is the probability density function (PDF) of the multivariate Gaussian distribution with mean $\bm{\upmu}_j$ and covariance $\mathbf{\Sigma}_j$, and these are calculated through EM algorithm.
For each Gaussian, the density of the features with a responsibility $\upgamma$ greater than $\phi=0.5$ (\ie, cluster) is measured. If the density of the fitted Gaussian is higher than $\psi=0.5$, it is determined that the Gaussian does not adequately reflect the boundaries that need to be disentangled; thus, a split is performed. The density is defined as follows:
\small
\begin{equation}
    \text{Density}_i = \frac{1}{N_i} \sum_{j=1}^{N_i} \|\mathbf{x}_{ij} - \frac{1}{N_i} \sum_{k=1}^{N_i} \mathbf{x}_{ik}\|_2
\end{equation}
\normalsize
where for cluster $i$, $\bx_{ij}$ is the $j$-th data point and $N_i$ is the number of data points.
In addition, to escape sub-optimal points while covering a large number of attributes, the split is also performed arbitrarily. During every split process, once a Gaussian is divided into two, it is re-optimized through the EM algorithm using the data belonging to each cluster.
After the split process, Gaussian filtering occurs to remove Gaussians that have too few data points in the cluster. This prevents small Gaussians from causing distortions in some features during feature alignment for new data.
Re-optimization also occurs after filtering, but since the split and filtering minimally alter the Gaussians, stability remains.

\ourparagraph{DyGA during the training process.}
Anchor selection is only possible when there is learned feature data. Therefore, after completing each epoch of training, we use the feature data to select anchors that will be used for future feature alignment. 
See the supplement for algorithmic details of the entire training process, including anchor selection and feature alignment.

\subsubsection{Feature Alignment}

Feature alignment refers to the process of shifting a feature $\mathbf{c} = [\mathbf{c}_1, \dots, \mathbf{c}_N]$ towards the mean $\bm{\upmu}_k$ of the Gaussian with the highest responsibility, as described in Eq.~\eqref{eq:responsibility}. Through feature alignment, the boundaries between clusters becomes definite. For $i \in [1, \dots, N]$ and aligned feature $\Tilde{\mathbf{c}}_i \in \mathbb{R}^D$, the feature alignment process is as follows:
\small
\begin{equation}
    \Tilde{\mathbf{c}}_i = \mathbf{c}_i + \delta (\bm{\upmu}_k - \mathbf{c}_i),
\label{eq:align}
\end{equation}
\normalsize
where $\delta = \lambda\exp\left(-\frac{1}{D} \sum_{j=1}^{D} \left| \frac{\mathbf{c}_{i}^j - \bm{\upmu}_{k}^j}{\mathbf{c}_{i}^j} \right| \right)$, $\lambda$ is a scale factor, and the superscript $j$ denotes the $j$-th element.
Since a feature located at the boundary between two Gaussians is sensitive, adjusting this feature is critical to the stability of the diffusion model training. Therefore, fully aligning it to the mean of either Gaussian or using too large $\lambda$ (\eg, $\lambda\geq1$) can negatively affect the overall training framework. This could cause the diffusion model to behave as if it were learning unconditional generation. Therefore, as described in Eq.~\eqref{eq:align}, the feature unit $\Tilde{\mathbf{c}}_{i}$ is an interpolation between the feature and the mean $\bm{\upmu}_{k}$ of the Gaussian. In this process, $\delta$ is determined by the distance between the feature and the mean. To prevent excessive variation due to the vector magnitude, $\delta$ is bounded by $\lambda$, considering the ratio of the difference to the feature. This ensures that the conditional diffusion model can stably utilize the output of the feature extractor, even when the distance from the Gaussian with the highest responsibility is too large. In this paper, $\lambda=0.1$ was used as the default.

\subsection{Skip Dropout}
\label{sec:skip}

Unlike VAE \cite{vae,beta-vae,anneal-vae,info-vae}, disentanglement diffusion models \cite{diff-ae,pdae,disdiff,fdae,encdiff} are structured with the junction of a feature extractor and the denoising network of a conditional diffusion model. In VAE, the decoder, which is the image generator, depends solely on the latent units. In contrast, the denoising network of the diffusion model relies on both the flow of the network according to the previous step's image $\mathbf{x}_t$ and the feature $\mathbf{c}$. 
Therefore, during the integrated training process of the diffusion denoising U-Net\cite{unet} and the feature extractor, it was necessary to prioritize the training of the feature extractor to achieve DRL-friendly training.
Considering this concern, we propose a skip dropout method inspired by DyLoRA \cite{dylora} as follows:
\begin{equation}
    \tilde{\mathbf{s}}_r = \mathbf{s}_r \odot \mathbf{m},
\end{equation}
where $\mathbf{m}_i \sim \text{Bernoulli}(p)$, $\mathbf{s}_r$ is the skip connection feature, and $\tilde{\mathbf{s}}_r$ is the dropout-applied skip connection feature.

DyLoRA, inspired by nested dropout\cite{nested-dropout}, uses higher dropout ratios for higher ranks so that the model relies more on lower-rank weights. This implies that dropout \cite{dropout} can concentrate information on specific weights. We aim to use this property to enhance the feature generation capability of the feature extractor without interfering with the training process of the denoising U-Net. The output of the feature extractor serves as the condition for the diffusion model, which is delivered as keys and values to specific blocks through cross attention. According to FreeU \cite{free-u} notation, this can be seen as part of the process of forming backbone features. Therefore, we adopt a method to drop out skip connection features at resolutions where conditions are not injected through cross attention, thereby emphasizing the learning of weights that create backbone features (\cref{fig:model}a). 

\ourparagraph{Remark.}
SD stochastically blinds some of the skip connection feature channels to prevent them from accumulating factor-specific information. 
As a result, the denoising U-Net yields the learning of the core image information to the feature extractor, which is not connected by skip connections. This allows the feature extractor to be sufficiently trained for disentangled representation.
\begin{table*}[t]
    \caption{Comparison with baselines on the FactorVAE score and DCI disentanglement metrics (mean $\pm$ std). \textbf{Bold} indicates the best, and \underline{underline} indicates the second-best.
    }
    \vspace{-10pt}
    \begin{center}
        \resizebox{\textwidth}{!}{
        \begin{tabular}{cccccccc}
            \toprule
            & \multirow{2}*{\textbf{Method}} & \multicolumn{2}{c}{Cars3D} & \multicolumn{2}{c}{Shapes3D} & \multicolumn{2}{c}{Mpi3D-toy} \\
            \cmidrule(lr){3-8}
            & & FactorVAE score$\uparrow$ & DCI$\uparrow$ & FactorVAE score$\uparrow$ & DCI$\uparrow$ & FactorVAE score$\uparrow$ & DCI$\uparrow$ \\
            \midrule
            \midrule
            \multirow{3}*{\textit{VAE}} & FactorVAE~\cite{shapes3d-factorvae} & $0.906 \pm 0.052$ & $0.161 \pm 0.019$ & $0.840 \pm 0.066$ & $0.611 \pm 0.082$ & $0.152 \pm 0.025$ & $0.240 \pm 0.051$ \\
            & $\beta$-TCVAE~\cite{mig-betaTCVAE} & $0.855 \pm 0.082$ & $0.140 \pm 0.019$ & $0.873 \pm 0.074$ & $0.613 \pm 0.114$ & $0.179 \pm 0.017$ & $0.237 \pm 0.056$ \\
            & DAVA~\cite{dava} & $0.940 \pm 0.010$ & $0.230 \pm 0.040$ & $0.820 \pm 0.030$ & $0.780 \pm 0.030$ & $0.480 \pm 0.050$ & $0.270 \pm 0.030$ \\
            \midrule
            \multirow{3}*{\textit{GAN}} & ClosedForm~\cite{CF} & $0.873 \pm 0.036$ & $0.243 \pm 0.048$ & $0.951 \pm 0.021$ & $0.525 \pm 0.078$ & $0.523 \pm 0.056$ & $0.318 \pm 0.014$  \\
            & GANSpace~\cite{GS} &$0.932 \pm 0.018$ & $0.209 \pm 0.031$ & $0.788 \pm 0.091$ & $0.284 \pm 0.034$ & $0.465 \pm 0.036$ & $0.229 \pm 0.042$ \\
            & DisCo-GAN~\cite{disco}& $0.855 \pm 0.074 $ & $0.271 \pm 0.037$ & $ 0.877 \pm 0.031 $ & $0.708 \pm 0.048 $ & $0.371 \pm 0.030 $ & $0.292 \pm 0.024$ \\
            \midrule
            \multirow{4}*{\textit{DM}} & DisDiff-VQ~\cite{disdiff} & $\boldsymbol{0.976} \pm 0.018$ & $0.232 \pm 0.019 $ & $ 0.902 \pm 0.043 $ & $0.723 \pm 0.013 $ & $0.617 \pm 0.070 $ & $0.337 \pm 0.057$ \\
            & FDAE~\cite{fdae} & $0.918 \pm 0.027$ & $0.232 \pm 0.418 $ & $ 0.987 \pm 0.023 $ & $0.917 \pm 0.038 $ & - & - \\
            & EncDiff~\cite{encdiff} & $0.773 \pm 0.060$ & $\underline{0.279} \pm 0.022 $ & $ \underline{0.999} \pm 0.000 $ & $\boldsymbol{0.969} \pm 0.030 $ & $\underline{0.872} \pm 0.049 $ & $\boldsymbol{0.685} \pm 0.044$ \\
            & \textbf{Ours} & $\underline{0.941} \pm 0.002$ & $\boldsymbol{0.414} \pm 0.013$ & $\boldsymbol{1.000} \pm 0.000$ & $\underline{0.938} \pm 0.001$ & $\boldsymbol{0.930} \pm 0.004$ & $\underline{0.627} \pm 0.002$ \\
            \bottomrule
        \end{tabular}
        }
    \end{center}
    \label{table:quant}
    \vspace{-10pt}
\end{table*}

\section{Experiments}
\label{sec:exp}

Our experiments aimed to verify how effective the proposed DyGA and SD are as inductive biases in the disentanglement of diffusion models. First, we introduce the setup including the dataset and metrics (\cref{sec:exp_setup}), and then validate the performance by comparing with state-of-the-art models (\cref{sec:quant}). Additionally, we examine the effectiveness of DyGA and SD in other diffusion-based disentanglement models (\cref{sec:fdae}). We also explore qualitative results through visualization (\cref{sec:visual}), and validate the performance of each methodology through an ablation study (\cref{sec:abla}). Subsequently, to evaluate the superiority of our representation, we assess it through a downstream task (\cref{sec:downsteam}). For more experiments and analyses, please refer to the supplementary material.

\subsection{Experimental Setup}
\label{sec:exp_setup}

\noindent\textbf{Datasets.}
We used the datasets Cars3D \cite{cars3d}, Shapes3D \cite{shapes3d}, and MPI3D-toy \cite{mpi3d}, which are commonly used in disentangled representation learning \cite{beta-vae,shapes3d-factorvae,disco,vct,disdiff,encdiff}. On the other hand, since these datasets do not include real-world data, we supplemented them with the CelebA dataset \cite{celeba}, which contains images of celebrity faces. To be consistent with previous research \cite{shapes3d-factorvae,mig-betaTCVAE,disdiff,encdiff}, we resized the images from the CelebA dataset to a resolution of 64$\times$64.

\vspace{3px}
\noindent\textbf{Implementation Details.}
We employed LDM \cite{ldm} as our diffusion model in order to reduce the computational cost while keeping in mind the application to high-resolution data. By using the commonly-utilized VQ-GAN, we mapped the pixel space image to a latent space. For image sampling, we used a 100-step DDIM sampler \cite{ddim}. To effectively extract features with a low computational burden, we used a simple CNN-based encoder similar to the one used in \cite{encdiff}. All experiments were conducted on a single NVIDIA RTX A6000. See supplement for more details.

\vspace{3px}
\noindent\textbf{Baselines.}
We compare our method with VAE-based methods—FactorVAE \cite{shapes3d-factorvae}, $\beta$-TCVAE \cite{mig-betaTCVAE}, and DAVA \cite{dava}—as well as GAN-based methods—
ClosedForm \cite{CF}, GANSpace \cite{GS}, and DisCo \cite{disco}. Furthermore, we include comparisons with diffusion models-based methods such as DisDiff \cite{disdiff}, FDAE \cite{fdae}, and EncDiff \cite{encdiff}.
For the CelebA dataset, we also compare against diffusion-based models that have attempted representation learning \cite{diff-ae,info-diff} as baselines.

\vspace{3px}
\noindent\textbf{Evaluation Metrics.}
To evaluate disentanglement performance, we employed the FactorVAE score\cite{shapes3d-factorvae} and DCI\cite{dci} as disentanglement metrics. This aims to understand overall performance by selecting metrics with relatively low covariance, based on a study analyzing the covariance between metrics\cite{challenging}.
For results for additional metrics—MIG\cite{mig-betaTCVAE}, Modularity\cite{modularity}, SAP score\cite{sap}, InfoMEC\cite{infomec-lq}, please refer to the supplementary material.
For the CelebA dataset, we measured disentanglement perfomance and image quality using TAD \cite{tad} and Fréchet Inception Distance (FID) \cite{fid}. Except for FID, all metrics are measured by reducing the dimensionality of vector-shaped latent units using PCA \cite{pca}, as in previous studies \cite{disco,disdiff,fdae,encdiff}.

\begin{table}[t]
\caption{Comparison of disentanglement and generation quality using the TAD and FID metrics (mean $\pm$ std) on CelebA dataset
}
    \vspace{-5pt}
    \resizebox{\linewidth}{!}{
    \begin{tabular}{lc@{\hspace{3em}}c}
        \toprule
        \multicolumn{1}{c}{Model} & TAD $\uparrow$ &  FID $\downarrow$ \\
        \midrule
        \midrule
        $\beta$-VAE \cite{beta-vae}	& $0.088 \pm 0.043$	& $99.8 \pm 2.4$ \\
        Diff-AE \cite{diff-ae}	& $0.155 \pm 0.010$	& $22.7 \pm 2.1$ \\
        InfoDiffusion \cite{info-diff} & $0.299 \pm 0.006$ & $23.6 \pm 1.3$  \\
        DisDiff \cite{disdiff} & $0.305 \pm 0.010$	& $18.2 \pm 2.1$ \\
        EncDiff \cite{encdiff} & $0.638 \pm 0.008$	& $14.8 \pm 2.3$ \\
        \midrule
        \textbf{Ours} & $\boldsymbol{0.954} \pm 0.024$	& $\boldsymbol{12.0} \pm 1.2$ \\
        \bottomrule
    \end{tabular}
    }
    \label{table:celeba}
\end{table}

\subsection{Comparison with the State-of-the-art Methods}
\label{sec:quant}

Here, we compare our proposed method with existing DRL works using quantitative measures.

\vspace{3px}
\noindent\textbf{Results: Cars3D, Shapes3D, and MPI3D-toy.}
As shown in \Cref{table:quant}, \textbf{1)} our method achieved a satisfactory FactorVAE score and the best DCI disentanglement performance on Cars3D. On Shapes3D and MPI3D-toy, it achieved either the best or second best results, and recorded a FactorVAE score of \textbf{0.930} on the MPI3D dataset, for which most baselines perform poorly.
\textbf{2)} EncDiff, which employs a simple CNN encoder within the LDM framework, can be considered our base model. Compared to EncDiff, we observed significant performance improvements on the Cars3D dataset, achieving comparable performance across all other datasets.

\begin{figure*}[t]
    \centering
    \includegraphics[width=\textwidth]{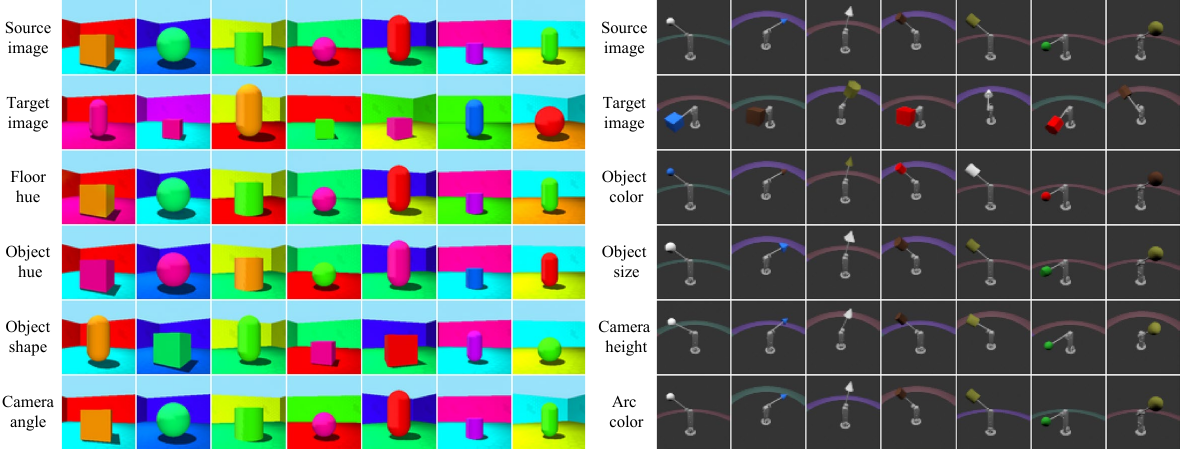}
    \caption{
    Latent interchange results.
    This figure shows the results of conditional generation using latent units as the condition, where a single latent unit of the source image is replaced with a latent unit from the target image. The first and second rows represent the source image and target image, respectively. The third to sixth rows show the source image with its attribute (\eg, \texttt{Floor hue}, \texttt{Camera angle}) changed to that of the target. (left) represents the Shapes3D dataset, while (right) represents the MPI3D dataset.
    }
    \vspace{-5pt}
    \label{fig:interpolation}
\end{figure*}

\vspace{3px}
\noindent\textbf{Results: CelebA.}
As shown in \Cref{table:celeba}, our method demonstrated a significant performance advantage over all other baselines in terms of TAD metric. This suggests that our modeling is not only effective on synthetic datasets designed for disentanglement (Cars3D, Shapes3D, and MPI3D), but also applicable to real-world datasets. Moreover, our method outperformed existing DM-based models in both TAD and FID with a TAD of \textbf{0.954}, suggesting that it overcomes the trade-off between image quality and disentanglement, which is a known limitation of VAE-based DRL.

\subsection{Application to Other Diffusion Models}
\label{sec:fdae}

Our methods can be plugged into other diffusion model-based DRL architectures where the diffusion models are conditioned on the outputs of any feature extractor. Through an example of combining with DisDiff\cite{disdiff} shown in \Cref{table:fdae}, we present that the proposed techniques are effective even when applied to other models.

Since EncDiff uses a LDM and a CNN-based encoder, it allows for an indirect comparison with our method, which is discussed in \cref{sec:quant}. To verify the effectiveness of our method on another diffusion model, we applied it to DisDiff, which has open source code available. We used the default hyperparameters and set the number of latent units $N=10$. DyGA was applied to the features passing through the DisDiff encoder, while SD was applied to the skip connection features used by the decoder. As a result, as shown in \Cref{table:fdae}, our method proved effective in other DM for DRL contexts as well. This demonstrates that our proposed DyGA and SD not only possess versatility across various models but also serve as effective inductive biases for disentanglement.

\subsection{Visualization Results}
\label{sec:visual}

\noindent\textbf{Latent Interchange Results.}
A straightforward way to prove whether the trained feature extractor outputs well-disentangled representations is by directly manipulating them. If the visualized results accurately reflect the intended changes in the representation, it means the latent units effectively capture the factors. Here, we refer to changing one of the latent units of a source image to that of a target image as latent interchange. The image generated using the modified representation as a condition reflects the single latent unit information from the target image onto the source image. As shown in \cref{fig:interpolation}, our method accurately conveys subtle differences, such as \texttt{camera angle}, \texttt{object color}, confirming its effectiveness in altering the image.

\begin{table}[t!]
    \caption{Validation of our method's effectiveness in DisDiff\cite{disdiff}}
    \vspace{-15pt}
    \centering
\renewcommand{\arraystretch}{1.3} 
\setlength{\tabcolsep}{3pt} 
    \begin{center}
        \resizebox{\linewidth}{!}{
        \begin{tabular}{lcccc}
            \toprule
            \multirow{2}*{\textbf{Method}} & \multicolumn{2}{c}{Shapes3D} & \multicolumn{2}{c}{MPI3D-toy}\\
            \cmidrule(lr){2-5}
            & \small FactorVAE score$\uparrow$ & DCI$\uparrow$ & \small FactorVAE score$\uparrow$ & DCI$\uparrow$ \\
            \midrule
            \midrule
            DisDiff & $ 0.902 \pm 0.043 $ & $0.723 \pm 0.013 $ & $ 0.617 \pm 0.070 $ & $0.337 \pm 0.057 $\\
            +DyGA, SD & $\boldsymbol{0.915} \pm 0.027$ & $\boldsymbol{0.745} \pm 0.018 $ & $ \boldsymbol{0.643} \pm 0.057 $ & $\boldsymbol{0.368} \pm 0.060 $\\
            \bottomrule
        \end{tabular}
        }
    \end{center}
    \label{table:fdae}
    \vspace{-10pt}
\end{table}

\begin{figure}[t!]
    \centering
    \includegraphics[width=1.0\linewidth]{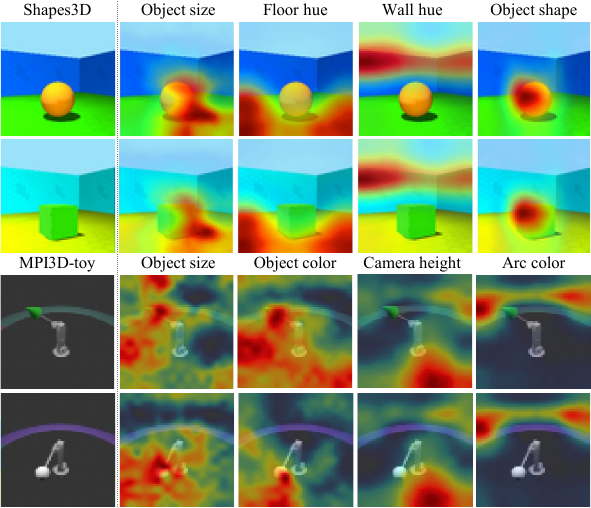}
    \vspace{-5pt}
    \caption{
    Attention map visualizations on Shapes3D and MPI3D-toy datasets. These results verify how well the image regions highlighted by the attention maps correspond to the factors represented by the latent units. In both datasets, the denoising U-Net focuses on the correct positions associated with the factors represented by the latent units (\eg, \texttt{object size}, \texttt{object shape}).
    }
    \vspace{-5pt}
    \label{fig:attention}
\end{figure}

\vspace{3px}
\noindent\textbf{Attention Map Visualization.}
The conditional diffusion model we used injects conditions through the cross-attention layer, allowing for visualization via the corresponding attention map. As shown in \cref{fig:attention}, this attention map shows which regions of the image the diffusion U-Net focuses on for noise prediction based on each latent unit. If a latent unit indeed represents a specific factor, the heat map will reflect the regions associated with that factor.
We visualized the average of all interpolated attention maps using the method from DAAM\cite{daam} in \cref{fig:attention} where it can be seen that the U-Net focuses on the factors represented by each latent unit in the Shapes3D dataset. Meanwhile, in the MPI3D-toy dataset, due to the variety of object orientations, the attention maps cover relatively broader areas, yet still capture the most defining features of the objects. For the \texttt{camera height}, the attention map shows a focus on the gap between the object and the floor as it changes with the camera angle, while also detecting the changes in the position of the arc.

\subsection{Ablation Study}
\label{sec:abla}

We further analyzed our proposed methods through an ablation study and discussed the results in \Cref{table:ablation}. First, we used a LDM with a CNN-based feature extractor as our base model and conducted experiments by applying DyGA and SD separately. The experiments were conducted using the MPI3D-toy dataset, which is sufficiently large and includes both simple elements like color and complex elements such as subtle differences in angle.
The experimental results showed that both DyGA and SD individually improved performance in terms of FactorVAE score and DCI disentanglement. When both methods were applied together, the best performance was achieved. Notably, as illustrated in \cref{fig:motivation}, we observed better attribute alignment within latent units when both methods were applied, and the clearer boundaries between attributes within latent units had a positive impact on performance, as we had anticipated.

\subsection{Downstream Tasks}
\label{sec:downsteam}

DRL is inherently focused on obtaining useful and meaningful representations within the machine learning domain. In this subsection, we evaluate how much the representation improves learning efficiency by using the gradient boosted trees (GBT). This involves a classification task using the representations, where the two kinds of accuracy $\text{Acc}_{1000}$ and $\text{Acc}_{100}$ obtained from training with 1000 and 100 representation samples, respectively, are compared to the accuracy $\text{Acc}$ obtained from training 10,000 samples. This serves as a metric to assess how efficient learning is by using fewer samples \cite{challenging}. To the best of our knowledge, there has been no prior attempts to perform this downstream task on the datasets we used. Therefore, we reproduced the existing models \cite{disdiff,fdae,encdiff} to serve as baselines, using the default hyperparameters suggested in the respective papers.
As shown in \Cref{table:down}, our method outperformed the existing baselines in the downstream tasks for both datasets. The significant improvement in the downstream tasks suggests that the inductive bias that separates attributes within latent units helps obtain better representations.

\begin{table}[t]
\centering
\caption{Ablation results on the MPI3D-toy dataset}
    \vspace{-10pt}
    \small
\scalebox{0.9}{
\renewcommand{\arraystretch}{1} 
\setlength{\tabcolsep}{14pt} 
    \begin{tabular}{lc@{\hspace{3em}}cc}
        \toprule
        \textbf{Method} & FactorVAE score$\uparrow$ & DCI$\uparrow$ \\
        \midrule
        \midrule
        Baseline & $0.856 \pm 0.004 $ & $0.586 \pm 0.002$ \\
        \midrule
        +DyGA  & $0.880 \pm 0.004$ & $0.626 \pm 0.001$ \\
        +SD  & $0.863 \pm 0.005$ & $0.615 \pm 0.002$ \\
        +DyGA, SD  & $\boldsymbol{0.930} \pm 0.004$ & $\boldsymbol{0.627} \pm 0.002$ \\
        \bottomrule
    \end{tabular}
    \label{table:ablation}
    }
\end{table}

\begin{table}[t!]
    \caption{Statistical efficiency for learning a GBT downstream task on Shapes3D and MPI3D-toy datasets}
    \vspace{-10pt}
    \centering
\renewcommand{\arraystretch}{1.3} 
\setlength{\tabcolsep}{3pt}
    \begin{center}
        \resizebox{\linewidth}{!}{
        \begin{tabular}{lcccc}
            \toprule
            \multirow{2}*{\textbf{Method}} & \multicolumn{2}{c}{Shapes3D} & \multicolumn{2}{c}{MPI3D-toy}\\
            \cmidrule(lr){2-5}
            & \small $\text{Acc}_{1000}/\text{Acc}$ & $\text{Acc}_{100}/\text{Acc}$ & \small $\text{Acc}_{1000}/\text{Acc}$ & $\text{Acc}_{100}/\text{Acc}$ \\
            \midrule
            \midrule
            DisDiff & $ 0.928 \pm 0.001 $ & $0.732 \pm 0.002 $ & $ 0.862 \pm 0.001 $ & $0.700 \pm 0.002 $\\
            FDAE & $ 0.979 \pm 0.002 $ & $0.751 \pm 0.019 $ & - & - \\
            EncDiff & $ 0.975 \pm 0.000 $ & $0.772 \pm 0.002 $ & $0.853 \pm 0.005 $ & $0.701 \pm 0.001$\\
            \midrule
            \textbf{Ours} & $\boldsymbol{0.990} \pm 0.000$ & $\boldsymbol{0.863} \pm 0.001 $ & $\boldsymbol{0.888} \pm 0.002$ & $\boldsymbol{0.757} \pm 0.001$ \\
            \bottomrule
        \end{tabular}
        }
    \end{center}
    \label{table:down}
\end{table}
\section{Conclusion}
\label{sec:conc}

In this study, we propose DyGA and SD, two novel approaches for disentangled representation learning by focusing directly on the latent unit itself.
Through empirical evaluation on various datasets, our proposed methods showcase remarkable performance in both disentanglement and downstream tasks.
Our DyGA, for the first time, introduces a disentanglement inductive bias which takes into account the attributes of each latent unit. This ensures that each latent unit is interpretable in real-world applications, enhancing their practicality. 
In addition, our proposed SD transforms the training framework of DMs for DRL into a DRL-friendly approach through stochastic blinding.
However, as the number of Gaussians in DyGA is not a continuous variable, the approach has limitations in handling attributes with continuous values.
Nonetheless, DyGA can potentially be applied to any dataset, since discrete variables can approximate continuous attributes.
In conclusion, our proposed methods emerge as powerful inductive biases for disentangled representation learning, ensuring the fidelity of latent units to reflect factors. 

\vspace{3px}
\noindent\textbf{Acknowledgments.} TThis work was supported in part by the IITP RS-2020-II201361 (AI Graduate School Program at Yonsei University), NRF RS-2024-00345806, and NRF RS-2023-00219019 funded by Korean Government (MSIT).

{\small
\bibliographystyle{ieee_fullname}
\bibliography{bib}
}

\clearpage
\onecolumn
\appendix
\section{Additional Materials}
An overview of the paper and a brief presentation video are available on our project page: \url{https://youngjun-jun.github.io/dis-dis-rep}.
\section{Datasets}

\begin{figure}[h]
    \centering
    \includegraphics[width=0.82\textwidth]{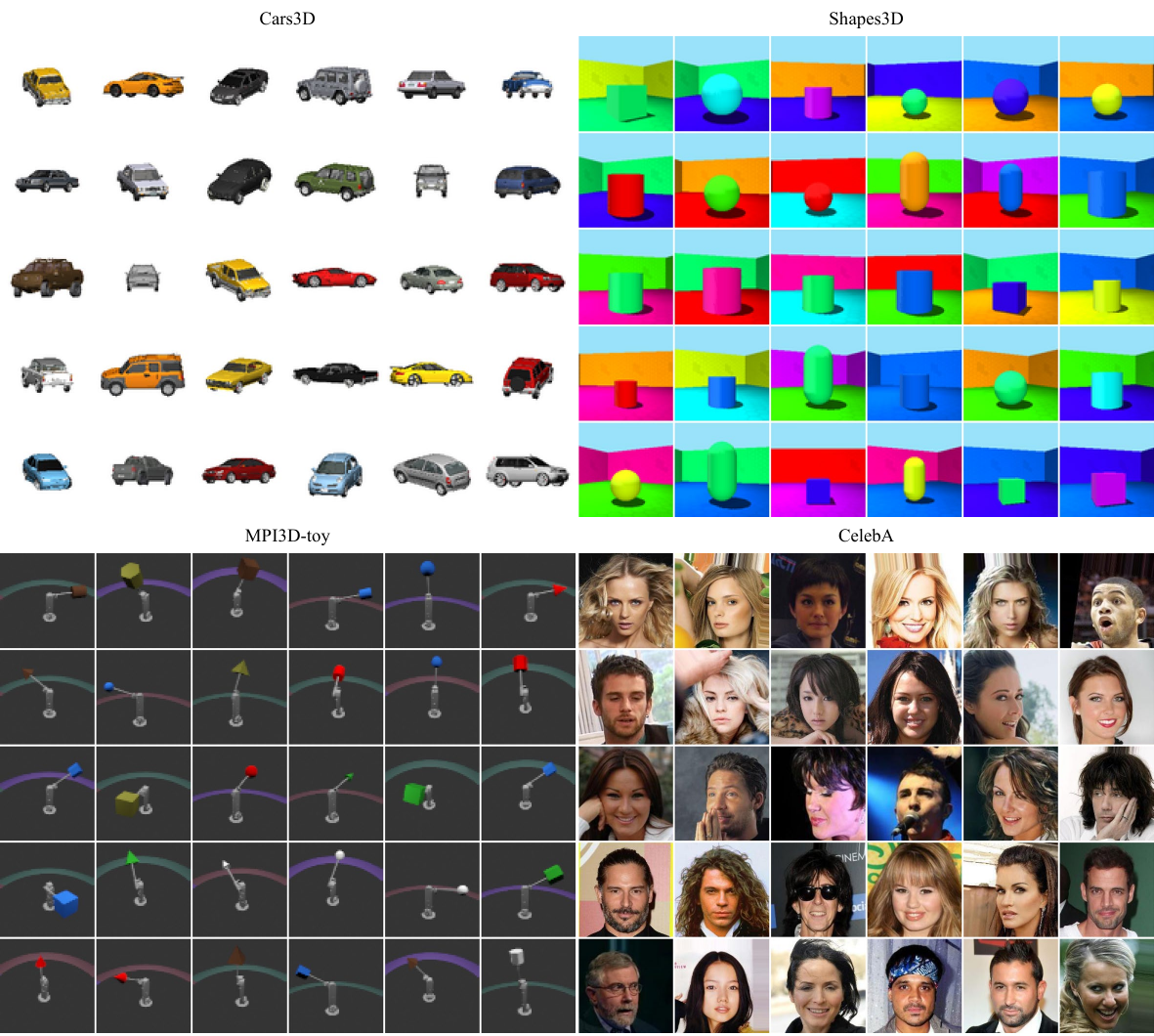}
    \caption{The samples from each dataset
    }
    \label{fig:dataset}
\end{figure}

This section provides an overview of the benchmark datasets (\cref{fig:dataset}).
The number of samples for each dataset is shown in \Cref{table:dataset}.

\vspace{3px}
\noindent\textbf{Cars3D\cite{cars3d}.} The Cars3D dataset is a car dataset created from CAD models with the following ground truth factors: 
elevation, azimuth, object type.

\vspace{3px}
\noindent\textbf{Shapes3D\cite{shapes3d}.} The Shapes3D dataset is composed of 3D shapes with the following ground truth factors:
floor hue, wall hue, object hue, scale, shape, orientation.

\vspace{3px}
\noindent\textbf{MPI3D-toy\cite{mpi3d}.} The MPI3D-toy dataset is part of the MPI3D dataset created to benchmark representation learning in simulated and real-world environments. It focuses on the toy type with the following ground truth factors: 
object color, object shape, object size, camera height, background color, first DOF, second DOF.

\vspace{3px}
\noindent\textbf{CelebA\cite{celeba}.} The CelebFaces Attributes Dataset (CelebA) is a face attributes dataset with 40 attributes. Although CelebA is not specifically designed for disentanglement, it allows for the validation of effects in real-world scenarios using various attributes.

\begin{table}[h]
\centering
\caption{The number of samples for each dataset}
    \fontsize{8}{10}\selectfont
    \begin{tabular}{cccc}
        \toprule
        Cars3D & Shapes3D & MPI3D & CelebA \\
        \midrule
        17,568  & 480,000 & 1,036,000 & 202,599 \\
        \bottomrule
    \end{tabular}
    \label{table:dataset}
\end{table}

\section{More Implementation Details}

\subsection{Training Details}

\subsubsection{Algorithms}

\noindent\textbf{Entire training framework.}
As in \cref{alg:framework}, the LDM training process using DyGA is conducted on an epoch-by-epoch basis. After every $r$ epochs, \textit{anchor selection} is performed for \textit{feature alignment} in the subsequent $r$ epochs. Considering the dataset size, we set $r=5$ for the Cars3D dataset, while $r=1$ was used for the others.

\vspace{3px}
\noindent\textbf{Anchor selection in DyGA.}
The anchor selection process (\cref{alg:fit}) consists of four stages. First, subspaces are updated using HDDC. Then, a Gaussian to split is selected based on density, and the two split Gaussians are updated using only the features within the cluster. Next, excessively small Gaussians that could act as outliers during feature alignment are removed. Finally, the Gaussians are updated using HDDC. These steps are performed at the latent unit level, with multi-processing utilized for each latent unit and accelerated through matrix multiplication. The anchor selection process for a single latent unit is completed within 30 seconds.

\vspace{3px}
\noindent\textbf{Feature alignment in DyGA.}
Na\"ively aligning features with the anchors selected through anchor selection hinders the backpropagation process. Therefore, as shown in \cref{alg:predict}, we use the Gumbel-softmax function \cite{gumbel-softmax} with a sufficiently small $\tau=0.0001$ to prevent distortion of the Gaussian mean while still allowing back-propagation. After this, the features undergo alignment in the direction of the anchor.

\begin{algorithm}[H]
    \caption{Entire training framework} \label{alg:framework} 
    \small
    \begin{algorithmic}
    \State \textbf{Input}: $\mathbf{X}$: dataset, $\{\sigma_t\}_{t=1}^T$: Noise schedule, $E(\cdot)$: VQ-encoder, $\mathbf{\epsilon_\theta}(\cdot)$: Denoising network, $f(\cdot)$: Feature extractor
    \For{$epoch = 1, \dotsc, max\_epoch$}
        \For{$\bx_0^i \in \mathbf{X}$}
            \State Sample $\mathbf{\epsilon} \sim \mathcal{N}(\mathbf{0}, \mathbf{I})$, $t \sim \mathrm{Uniform}(\{0, 1, \dots, T\})$
            \State $\bz_t^i \leftarrow E(\bx_0^i)+\sigma_t \mathbf{\epsilon}$
            \State $\mathbf{c}_i \leftarrow f(\bx_0^i)$
            \If{$epoch \geq r$}
                $\Tilde{\mathbf{c}}_i \leftarrow Feature\_alignment(\mathbf{c}_i)$
            \Else
                $\;\Tilde{\mathbf{c}}_i \leftarrow \mathbf{c}_i$
            \EndIf 
            \State Predict noise $\mathbf{\epsilon}_\theta(\bz_t, t, \tilde{\mathbf{c}}_i)$
            \State Compute loss and gradient
            \State Update parameters $\theta$
        \State \textbf{Anchor\_selection}
        \EndFor
    \EndFor
    \end{algorithmic}
\end{algorithm}
\clearpage
\begin{multicols}{2}
\begin{algorithm}[H]
    \caption{Anchor selection of the $k^{th}$ cluster} \label{alg:fit} 
    \small
    \begin{algorithmic}
    \State \textbf{Input}: $\mathbf{c} \in \mathbb{R}^{d}$: latent units
    \State \textbf{Output}: $\mu_k$, $\Sigma_k$, $w_k$, $\Lambda_k$, $\mathbf{v}_k$, $D_k$

    \State \textbf{Initialize} $\mu_k$, $\Sigma_k$, $w_k$, $\Lambda_k$, $\mathbf{v}_k$, $D_k$
    \For{$iter = 1, \dotsc, max\_iter$}
        \State \textbf{E-step}: Calculate $r_{ik}$
        \State \textbf{M-step}: Update $\mu_k$, $\Sigma_k$, $w_k$
        \State \textbf{Update subspaces}: Compute $\Lambda_k$, $\mathbf{v}_k$, $D_k$
    \EndFor

    \State \textbf{Calculate density} of $clusters \in \mathcal{F}$
    \While{$\mathcal{F} \neq \emptyset$}
        \State \textbf{Choose cluster} $\mathbf{f} \in \mathcal{F}$
        \State \textbf{Split cluster} $\mathbf{f}$
        \State \textbf{E-step}, \textbf{M-step}, and \textbf{Update subspaces} in the $cluster$
        \State \textbf{Update $\mathcal{F}$}
        \EndWhile
    \State \textbf{Remove small Gaussians}
    \For{$iter = 1, \dotsc, max\_iter$}
        \State \textbf{E-step}, \textbf{M-step}, and \textbf{Update subspaces}
    \EndFor
    
    \end{algorithmic}
\end{algorithm}

\begin{algorithm}[H]
    \caption{Feature alignment} \label{alg:predict} 
    \small
    \begin{algorithmic}
    \State \textbf{Input}: $\mathbf{c} \in \mathbb{R}^{d}$: latent unit, $\tau$: softmax parameter, $\lambda$: feature alignment parameter
    \State \textbf{Output}: $\hat{\mathbf{c}} \in \mathbb{R}^{d}$

    \State \textbf{E-step}: Calculate responsibility $r$
    \State \textbf{Gumbel-Softmax}:
    \begin{enumerate}
        \item Generate Gumbel noise $g_{k}$
        \item Compute $y_{k} = \text{softmax}\left(\frac{r_{k} + g_{k}}{\tau}\right)$
    \end{enumerate}

    \State \textbf{Update} $\upmu_k \leftarrow \sum_{k} y_{k} \upmu_k$
    \State \textbf{Compute multiplier} $\delta \leftarrow \lambda\exp\left(-\frac{1}{d} \sum_{j=1}^{d} \left| \frac{\mathbf{c}_{i}^j - \boldsymbol{\upmu}_{k}^j}{\mathbf{c}^j} \right| \right)$
    \State \textbf{Align latent unit} $\Tilde{\mathbf{c}} \leftarrow \mathbf{c} + \delta (\boldsymbol{\upmu}_k - \mathbf{c})$
    \end{algorithmic}
\end{algorithm}
\end{multicols}

\subsubsection{LDM training details}
During the training process of LDM, we set the batch size to 512 for the Cars3D\cite{cars3d}, Shapes3D\cite{shapes3d}, and MPI3D-toy\cite{mpi3d} datasets, and 64 for the CelebA dataset. The learning rate and Exponential Moving Average (EMA) rate were set to 0.0001 and 0.9999, respectively, for all datasets, following EncDiff\cite{encdiff}.

\subsection{Model Architecture}
The latent diffusion model (LDM)\cite{ldm} we used is a diffusion model for images in the latent space reduced by VQ-GAN\cite{vq-gan}. The architectures of VQ-GAN and the LDM denoising U-Net are shown in \Cref{table:vqgan} and \Cref{table:ldm-unet}, respectively. Additionally, the feature extractor (\Cref{table:feature-extractor}), similar to previous studies \cite{fdae,encdiff}, uses a structure where a scalar value is taken for each latent unit through a CNN module and then passed through independent MLP modules. We used the same settings for all datasets. 

\begin{multicols}{2}
    \begin{table}[H]
    \centering
    \caption{VQ-GAN architecture parameters}
    \begin{tabular}{l@{\hspace{3em}}c}
        \toprule
        Parameter & Value \\
        \midrule
        Embedding dimensionality & 3 \\
        Number of embeddings & 2048 \\
        Channels in first conv layer & 32 \\
        Channel multipliers & [1, 2, 4] \\
        Residual blocks per layer & 2 \\
        Dropout rate & 0.0 \\
        Discriminator start epoch & 0 \\
        Discriminator loss weight & 0.75 \\
        Codebook loss weight & 1.0 \\
        \bottomrule
    \end{tabular}
    \label{table:vqgan}
    \end{table}
    
    \begin{table}[H]
    \centering
    \caption{Denoising U-Net architecture parameters}
    \begin{tabular}{l@{\hspace{3em}}c}
        \toprule
        Parameter & Value \\
        \midrule
        Input image size & 16 \\
        Input channels & 3 \\
        Base channels & 64 \\
        Attention resolutions & [1, 2, 4] \\
        Residual blocks per layer & 2 \\
        Channel multipliers & [1, 2, 4, 4] \\
        Attention heads & 8 \\
        Scale-shift normalization & True \\
        Context dimension & 32 \\
        Dropout rate & 0.1 \\
        Skip dropout rate & 0.2 \\
        Noise schedule & Linear \\
        Diffusion timestep & 1000 \\
        \bottomrule
    \end{tabular}
    \label{table:ldm-unet}
    \end{table}
\end{multicols}

\begin{table}[H]
\centering
\caption{Feature extractor architecture}
\resizebox{0.45\textwidth}{!}{
\begin{tabular}{l@{\hspace{3em}}c}
    \toprule
    CNN module \\
    \midrule
    Conv $7\times7\times3\times64$, stride$=1$, padding$=3$ \\
    BatchNorm \\
    ReLU \\
    Conv $4\times4\times64\times128$, stride$=1$, padding$=3$ \\
    BatchNorm \\
    ReLU \\
    Conv $4\times4\times128\times256$, stride$=2$, padding$=1$ \\
    BatchNorm \\
    ReLU \\
    Conv $4\times4\times256\times256$, stride$=2$, padding$=1$ \\
    BatchNorm \\
    ReLU \\
    Conv $4\times4\times256\times256$, stride$=2$, padding$=1$ \\
    BatchNorm \\
    ReLU \\
    FC $4096\times4096$ \\
    ReLU \\
    FC $4096\times256$ \\
    ReLU \\
    FC $256\times N$ \\
    \bottomrule
    \toprule
    MLP module \\
    \midrule
    FC $1\times256$ \\
    ReLU \\
    FC $256\times512$ \\
    ReLU \\
    FC $512\times32$ \\
    ReLU \\
    \bottomrule
\end{tabular}
}
\label{table:feature-extractor}
\end{table}

\section{More Experiments}

\subsection{Results for more metrics}

In this subsection, as shown in \Cref{table:metrics}, we evaluate the disentanglement performance of our method using not only the FactorVAE score \cite{shapes3d-factorvae} and DCI disentanglement \cite{dci}, but also additional metrics such as MIG \cite{mig-betaTCVAE}, Modularity score \cite{modularity}, SAP score \cite{sap}, and InfoMEC~InfoM score \cite{infomec-lq}.

\begin{table}[h]
\centering
\caption{Results for more metrics on Cars3D, Shapes3D and MPI3D-toy datasets}
    \small
    \begin{tabular}{l@{\hspace{2em}}cccccc}
        \toprule
        \textbf{Dataset} & FactorVAE score$\uparrow$ & DCI$\uparrow$ & MIG$\uparrow$ & Modularity score$\uparrow$ & SAP score$\uparrow$ & InfoM score$\uparrow$ \\
        \midrule
        \midrule
        Cars3D  & $0.941 \pm 0.002$ & $0.414 \pm 0.013$ & $0.109 \pm 0.002$ & $0.934 \pm 0.001$ & $0.009 \pm 0.002$ & $0.417 \pm 0.004$ \\
        \midrule
        Shapes3D  & $1.000 \pm 0.000$ & $0.938 \pm 0.001$ & $0.507 \pm 0.002$ & $0.930 \pm 0.001$ & $0.183 \pm 0.006$ & $0.569 \pm 0.002$ \\
        \midrule
        MPI3D-toy  & $0.930 \pm 0.004$ & $0.627 \pm 0.002$ & $0.364 \pm 0.001$ & $0.882 \pm 0.002$ & $0.174 \pm 0.002$ & $0.495 \pm 0.003$ \\
        \bottomrule
    \end{tabular}
    \label{table:metrics}
\end{table}

\subsection{DyGA and SD Analysis}

\begin{table}[h]
\centering
\caption{Comparison between DyGA and FSQ}
    \small
    \begin{tabular}{l@{\hspace{2em}}cc}
        \toprule
        \textbf{Method} & FactorVAE score$\uparrow$ & DCI$\uparrow$ \\
        \midrule
        \midrule
        Baseline & $0.856 \pm 0.004 $ & $0.586 \pm 0.002$ \\
        \midrule
        +FSQ, SD  & $0.606 \pm 0.006$ & $0.384 \pm 0.002$ \\
        +DyGA, SD  & $\boldsymbol{0.930} \pm 0.004$ & $\boldsymbol{0.627} \pm 0.002$ \\
        \bottomrule
    \end{tabular}
    \label{table:fsq}
\end{table}

\noindent\textbf{DyGA Analysis.}
Our proposed \textit{Dynamic Gaussian Anchoring} (DyGA) can dynamically adjust the position and number of anchors. A similar approach can be found in the quantization methods used in variational autoencoder (VAE)\cite{vae}-based approaches \cite{infomec-lq,tripod}. One such method, finite scalar quantization (FSQ)\cite{fsq}, can be applied to diffusion-based models as a way to organize latent units. However, when we set the number of codebooks to 1 and the levels to [8, 5, 5], performance actually degraded, as shown in Table 6, with similar performance drops observed in other settings. This suggests that attempts to organize latent units in diffusion-based models using a fixed number of quantization values or fixed quantized values may not be suitable. Therefore, we introduced DyGA, which dynamically adjusts the number and positions of anchors, resulting in significant performance improvements.

\begin{table}[h]
\centering
\caption{Skip and Backbone Dropout}
    \small
    \begin{tabular}{l@{\hspace{2em}}cc}
        \toprule
        \textbf{Method} & FactorVAE score$\uparrow$ & DCI$\uparrow$ \\
        \midrule
        \midrule
        +BD-0.1  & $0.798 \pm 0.005$ & $0.610 \pm 0.002$ \\
        \midrule
        Baseline & $0.856 \pm 0.004 $ & $0.586 \pm 0.002$ \\
        \midrule
        +SD-0.1  & $0.852 \pm 0.005$ & $0.598 \pm 0.007$ \\
        +SD-0.2  & $\boldsymbol{0.863} \pm 0.005$ & $\boldsymbol{0.615} \pm 0.002$ \\
        +SD-0.3  & $0.855 \pm 0.006$ & $0.603 \pm 0.002$ \\
        \bottomrule
    \end{tabular}
    \label{table:sd}
\end{table}

\noindent\textbf{SD Analysis.}
\textit{Skip Dropout} (SD) stochastically blinds skip connection features to prevent the continuous accumulation of factor information in specific weights. This forces the diffusion U-Net to rely on the backbone features connected to the feature extractor, which continuously provide factor information. To analyze this effect more deeply, we examine the impact of SD by using backbone dropout (BD), which dropouts backbone features in parts with skip connections. 
On the other hand, it has been shown that dropout rates between 0.4 and 0.8 do not affect performance improvement \cite{dropout}, and we have experimentally found that a high skip dropout ratio (\eg, $1-p=1.0$) negatively impacts the convergence of the diffusion model. Therefore, we used skip dropout rates of 0.1, 0.2, and 0.3, while the backbone dropout rate was set to 0.1. 
As a result, as shown in \Cref{table:sd}, an appropriate skip dropout rate was effective, whereas backbone dropout degraded performance.

\subsection{Latent Interpolation}

\begin{figure*}[b]
    \centering
    \includegraphics[width=\textwidth]{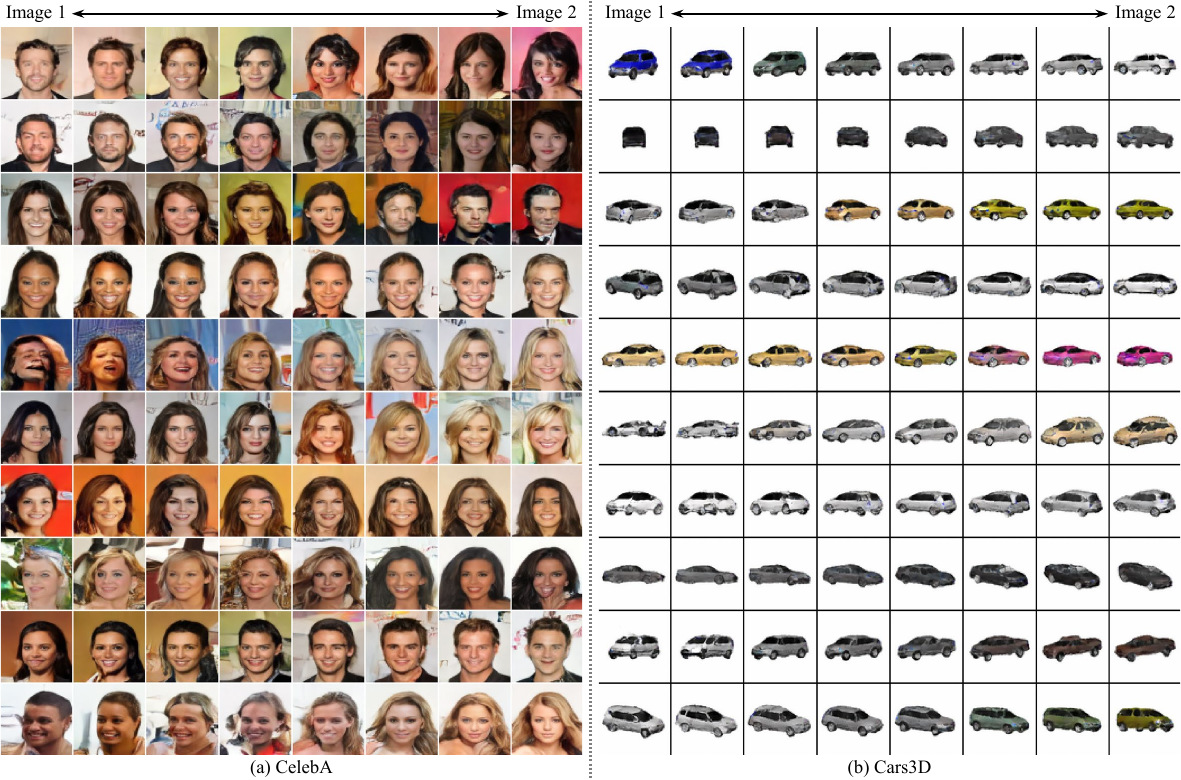}
    \caption{
    Visualization of latent interpolation on the Cars3D and CelebA datasets. \textbf{(a)} For CelebA, we observe natural transitions between two images in terms of \texttt{hair color}, \texttt{hair style}, \texttt{skin color}, \texttt{background color}, \texttt{gender}, and \texttt{smile}. \textbf{(b)} Similarly, in Cars3D, we observe smooth changes in \texttt{vehicle type}, \texttt{color}, \texttt{azimuth}, and \texttt{elevation}.
    }
    \label{fig:latent-interp}
\end{figure*}

There has been no exploration of the disentangled representation space of diffusion-based models (\ie, the space of outputs from the feature extractor). In this subsection, we visualize images generated by interpolating latent units extracted from two images. This intermediate image generation enables image morphing\cite{morphing-1,morphing-2,morphing-3,diffmorpher}, which shows the transformation between images as a video. Experimentally, when using spherical linear interpolation (slerp) \cite{slerp} for the diffusion latent interpolation method in DiffMorpher\cite{diffmorpher}, the generated intermediate images appeared unnatural. Instead, as shown in \cref{fig:latent-interp}, images generated using linear interpolation suggest the potential for image morphing through disentanglement.

\section{More Visualizations}

\subsection{Training loss curve}

\begin{figure}[b]
    \centering
    \includegraphics[width=0.8\textwidth]{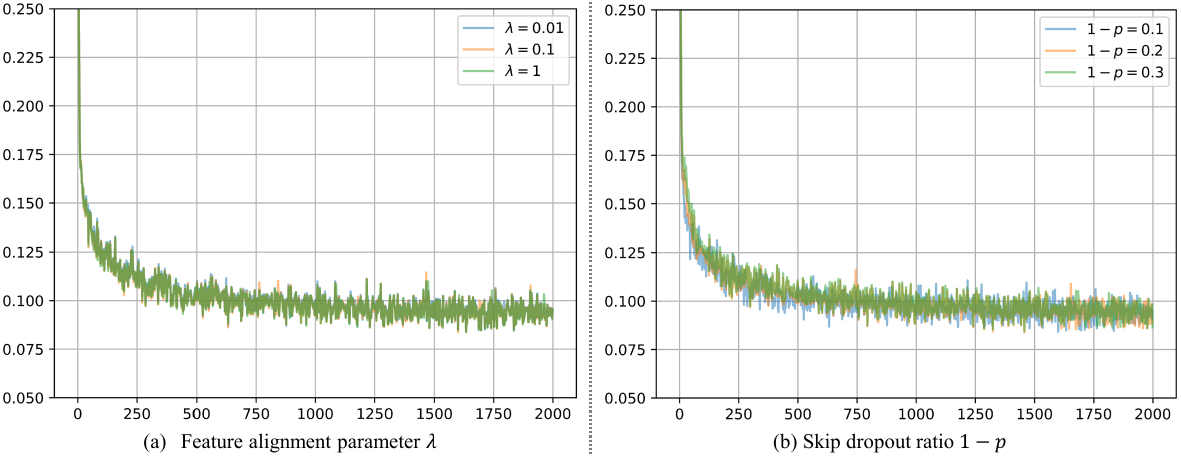}
    \caption{
    Training loss curve
    }
    \label{fig:loss}
\end{figure}

In this subsection, we plot the training loss curve according to different parameters. From \cref{fig:loss}, we can confirm that our methods remain stable despite changes in training parameters ($\lambda$: feature alignment parameter, $1-p$: skip dropout ratio).

\subsection{More Latent Interchange Results}

\begin{figure*}[t]
    \centering
    \includegraphics[width=\textwidth]{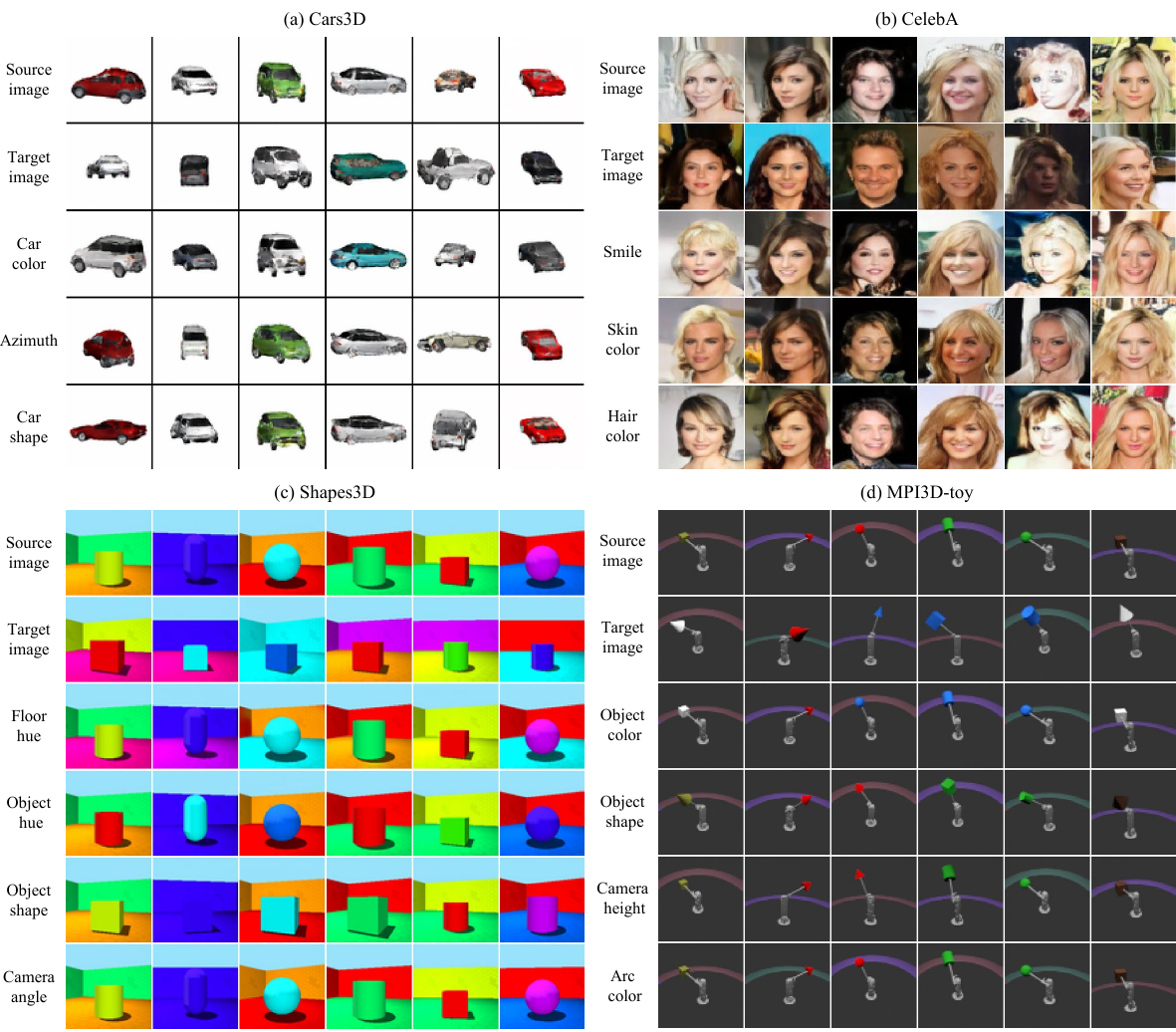}
    \caption{
    Latent interchange results on Cars3D, Shapes3D, MPI3D-toy, and CelebA datasets
    }
    \label{fig:more-inter}
\end{figure*}

One way to verify if a trained feature extractor extracts well-disentangled representations is to manipulate them directly. If the visualized results accurately reflect the intended changes in representation, it indicates that the latent units faithfully represent the factors. Here, we refer to changing one of the latent units of the source image to the latent unit of the target image as a latent interchange. The images generated using the latent units created through latent interchange conditionally alter the source image using the single latent unit information of the target image. \cref{fig:more-inter} visualizes how well the feature extractor is trained in each dataset through latent interchange. Our method demonstrates that the images generated using latent interchange effectively reflect a single characteristic of the target image.

\subsection{More Attention Map Visualizations}

\begin{figure}[t!]
    \centering
    \includegraphics[width=0.8\linewidth]{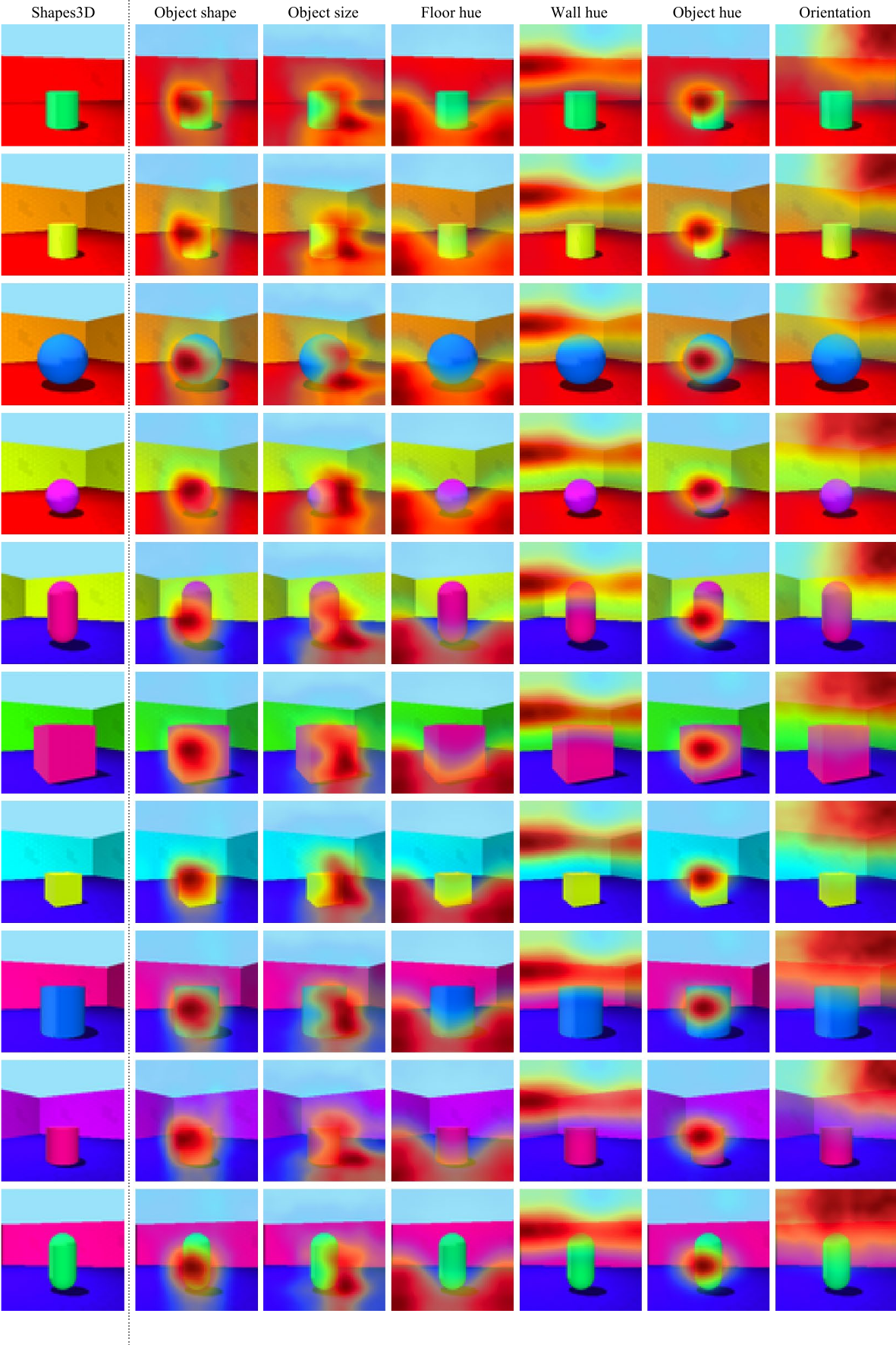}
    \caption{
    Attention map visualizations on Shapes3D dataset
    }
    \label{fig:more-attention1}
\end{figure}

\begin{figure}[t!]
    \centering
    \includegraphics[width=1.0\linewidth]{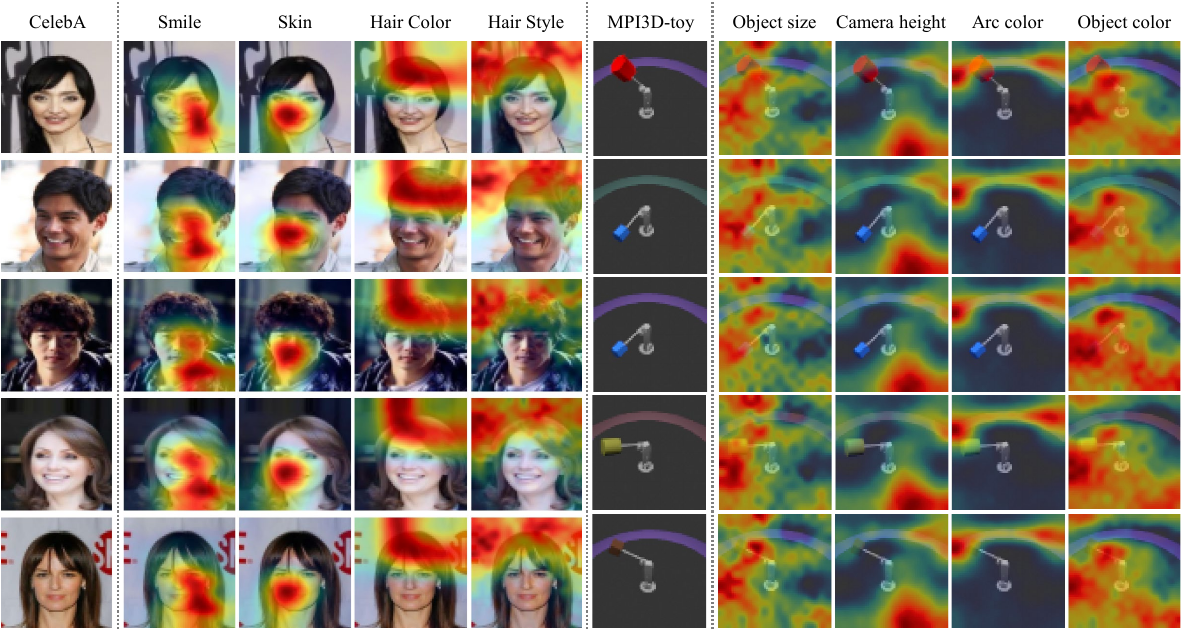}
    \caption{
    Attention map visualizations on CelebA and MPI3D-toy datasets
    }
    \label{fig:more-attention2}
\end{figure}

Since our diffusion model receives conditions through cross attention, it is possible to visualize the attention map. This attention map shows which areas of the image are being highlighted, indicating which parts of the image each latent unit uses to generate. The results on various datasets can be seen in \cref{fig:more-attention1,fig:more-attention2}.

\subsection{More Latent Unit Visualizations}

\begin{figure*}[h]
    \centering
    \includegraphics[width=\textwidth]{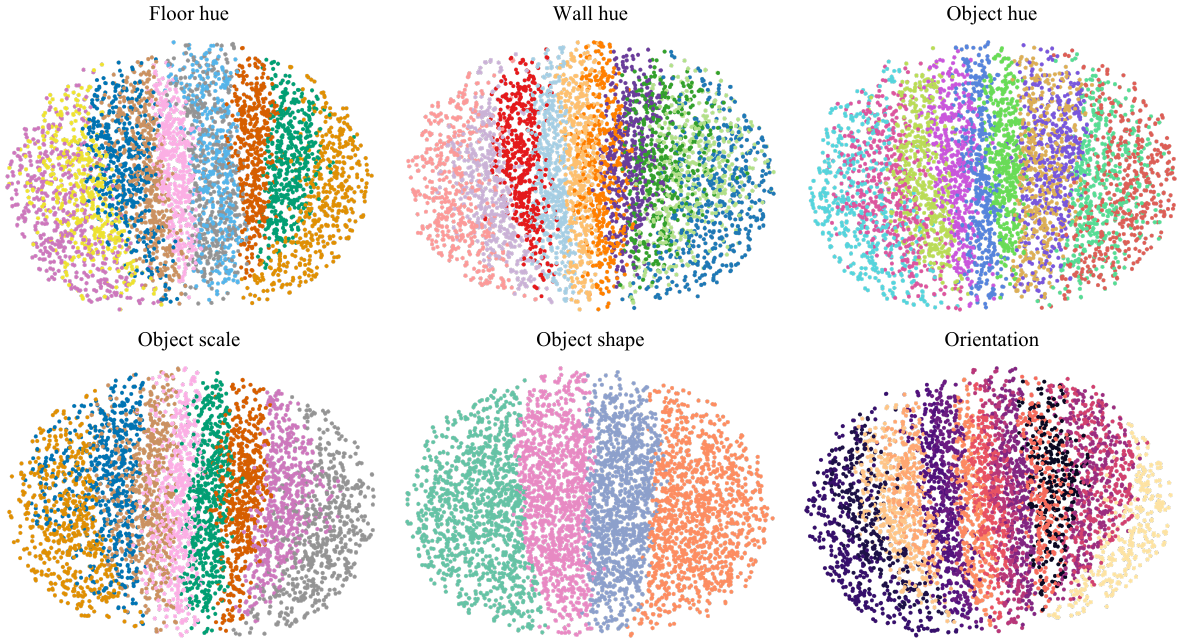}
    \caption{
    Visualization of latent units on Shapes3D dataset.
    Each visualized latent unit here is associated with the written factor (\eg, it has the highest normalized mutual information). Each color represents an attribute of the factor, such as \texttt{red} in object color or \texttt{cube} in object shape.
    }
    \label{fig:more-latent}
\end{figure*}

Disentangled representation learning aims to make each latent unit sensitive to a single fundamental factor while being invariant to the other factors. At this point, the latent unit with the highest association (\eg, normalized mutual information) to a given factor best reflects the information of that factor. Therefore, we visualize the latent units for various data points in the Shapes3D dataset (\cref{fig:more-latent}) to see how well the latent units separate the factor attributes.

\end{document}